\documentclass[journal]{IEEEtran}
\usepackage{cite}
\usepackage{amsmath,amssymb,amsfonts}
\usepackage{algorithmic}
\usepackage{graphicx}
\usepackage{tabularx}
\usepackage{textcomp}
\usepackage{caption}
\usepackage{subcaption}
\usepackage{xcolor}
\usepackage[
   separate-uncertainty = true,
   multi-part-units =repeat
]{siunitx}

\newcommand{\Lagr}{\mathcal{L}}
\definecolor{normalGreen}{RGB}{141, 208, 63}

\begin{document}
%
\title{Regressing Word and Sentence Embeddings\\for Regularization of Neural Machine Translation}
%
%
%

\author{Inigo~Jauregi~Unanue,~Ehsan~Zare~Borzeshi,~and~Massimo~Piccardi, \textit{Senior Member}, \textit{IEEE}
\thanks{I. Jauregi Unanue is with University of Technology Sydney, Sydney, NSW 2007, Australia, and the RoZetta Institute, Sydney, NSW 2000, Australia, e-mail: \tt inigo.jauregi@rozettainstitute.com}

\thanks{E. Zare Borzeshi is with Microsoft Commercial Software Engineering (CSE), Sydney, NSW 2113, Australia, e-mail: \tt ehsan.zareborzeshi@microsoft.com}
\thanks{M. Piccardi is with University of Technology Sydney, Sydney, NSW 2007, Australia, e-mail: \tt massimo.piccardi@uts.edu.au}}

%
%

\markboth{}{}
%



\maketitle


\begin{abstract}
In recent years, neural machine translation (NMT) has become the dominant approach in automated translation. However, like many other deep learning approaches, NMT suffers from overfitting when the amount of training data is limited. This is a serious issue for low-resource language pairs and many specialized translation domains that are inherently limited in the amount of available supervised data. For this reason, in this paper we propose regressing word (ReWE) and sentence (ReSE) embeddings at training time as a way to regularize NMT models and improve their generalization. During training, our models are trained to jointly predict categorical (words in the vocabulary) and continuous (word and sentence embeddings) outputs. An extensive set of experiments over four language pairs of variable training set size has showed that ReWE and ReSE can outperform strong state-of-the-art baseline models, with an improvement that is larger for smaller training sets (e.g., up to $+5.15$ BLEU points in Basque-English translation). Visualizations of the decoder's output space show that the proposed regularizers improve the clustering of unique words, facilitating correct predictions. In a final experiment on unsupervised NMT, we show that ReWE and ReSE are also able to improve the quality of machine translation when no parallel data are available.
\end{abstract}


%
\IEEEpeerreviewmaketitle

\section{Introduction}
\label{introduction}

Machine translation (MT) is a field of natural language processing (NLP) focussing on the automatic translation of sentences from a \textit{source} language to a \textit{target} language. In recent years, the field has been progressing quickly mainly thanks to the advances in deep learning and the advent of neural machine translation (NMT). The first NMT model was presented in 2014 by Sutskever et al. \cite{sutskever2014} and consisted of a plain encoder-decoder architecture based on recurrent neural networks (RNNs). In the following years, a series of improvements has led to major performance increases, including the attention mechanism (a word-aligment model between words in the source and target sentences) \cite{bahdanau2014,luong2015} and the transformer (a non-recurrent neural network that offers an alternative to RNNs and makes NMT highly parallelizable)  \cite{vaswani2017attention}. As a result, NMT models have rapidly outperformed traditional approaches such as phrase-based statistical machine translation (PBSMT) \cite{koehn2007moses} in challenging translation contexts (e.g., the WMT conference series). Nowadays, the majority of commercial MT systems utilise NMT in some form.

However, NMT systems are not exempt from limitations. The main is their tendence to overfit the training set due to their large number of parameters. This issue is common to many other tasks that use deep learning models and it is caused to a large extent by the way these models are trained: maximum likelihood estimation (MLE). As pointed out by Elbayad et al. \cite{elbayad2018token}, in the case of machine translation, MLE has two clear shortcomings that contribute to overfitting:

\begin{enumerate}
	\item \textbf{Single ground-truth reference}: Usually, NMT models are trained with translation examples that have a single reference translation in the target language. MLE tries to give all the probability to the words of the ground-truth reference and zero to all others. Nevertheless, a translation that uses different words from the reference (e.g. paraphrase sentences, synonyms) can be equally correct. Standard MLE training is not able to leverage this type of information since it treats every word other than the ground truth as completely incorrect.
	\item \textbf{Exposure bias}\cite{bengio2015scheduled}: NMT models are trained with ``teacher forcing'', which means that the previous word from the reference sentence is given as input to the decoder for the prediction of the next. This is done to speed up training convergence and avoid prediction drift. However, at test time, due to the fact that the reference is not available, the model has to rely on its own predictions and the performance can be drastically lower. 
\end{enumerate}

Both these limitations can be mitigated with sufficient training data. In theory, MLE could achieve optimal performance with infinite training data, but in practice this is impossible as the available resources are always limited. In particular, when the training data are scarce such as in low-resource language pairs or specific translation domains, NMT models display a modest performance, and other traditional approaches (e.g., PBSMT)\cite{koehn2017six} often obtain better accuracies. As such, generalization of NMT systems still calls for significant improvement.

In our recent work~\cite{jauregi2019rewe}, we have proposed a novel regularization technique that is based on co-predicting words and their embeddings (``regressing word embeddings'', or ReWE for short). ReWE is a module added to the decoder of a sequence-to-sequence model so that, during training, the model is trained to jointly predict the next word in the translation (categorical value) and its pre-trained word embedding (continuous value). This approach can leverage the contextual information embedded in pre-trained word vectors to achieve more accurate translations at test time. ReWE has been showed to be very effective over low/medium size training sets~\cite{jauregi2019rewe}. In this paper, we extend this idea to its natural counterpart: sentence embedding. We propose regressing sentence embeddings (ReSE) as an additional regularization method to further improve the accuracy of the translations. ReSE uses a self-attention mechanism to infer a fixed-dimensional sentence vector for the target sentence. During training, the model is trained to regress this inferred vector towards the pre-trained sentence embedding of the ground-truth sentence. The main contributions of this paper are:

\begin{itemize}
	\item The proposal of a new regularization technique for NMT based on sentence embeddings (ReSE).
	\item Extensive experimentation over four language pairs of different dataset sizes (from small to large) with both word and sentence regularization. We show that using both ReWE and ReSE can outperform strong state-of-the-art baselines based on long short-term memory networks (LSTMs) and transformers.
	\item Insights on how ReWE and ReSE help to improve NMT models. Our analysis shows that these regularizers improve the organization of the decoder's output vector space, likely facilitating correct word classification.
	\item Further experimentation of the regularizer on \textit{unsupervised} machine translation, showing that it can improve the quality of the translations even in the absence of parallel training data.
\end{itemize}

The rest of this paper is organized as follows. Section \ref{related_work} presents and discusses the related work. Section \ref{baseline_nmt_model} describes the model used as baseline while Section \ref{rewe_and_rese} presents the proposed regularization techniques, ReWE and ReSE. Section \ref{experiments} describes the experiments and analyzes the experimental results. Finally, Section \ref{conclusion} concludes the paper.

\section{Related Work}
\label{related_work}

The related work is organized over the three main research subareas that have motivated this work: \textit{regularization techniques}, \textit{word and sentence embeddings} and \textit{unsupervised NMT}.

\subsection{Regularization Techniques}
\label{subsec_reg}

In recent years, the research community has dedicated much attention to the problem of overfitting in deep neural models. Several regularization approaches have been proposed in turn such as dropout \cite{srivastava2014dropout,gal2016theoreticallys}, data augmentation \cite{fadaee2017data} and multi-task learning \cite{gu2018meta,clark2018semi}. Their common aim is to encourage the model to learn parameters that allow for better generalization.

In NMT, too, mitigating overfitting has been the focus of much research. As mentioned above, the two, main acknowledged problems are the single ground-truth reference and the exposure bias. For the former, Fadee et al. \cite{fadaee2017data} have proposed augmenting the training data with synthetically-generated sentence pairs containing rare words. The intuition is that the model will be able to see the vocabulary's words in more varied contexts during training. Kudo \cite{kudo2018subword} has proposed using variable word segmentations to improve the model's robustness,  achieving notable improvements in low-resource languages and out-of-domain settings. Another line of work has focused on ``smoothing'' the output probability distribution over the target vocabulary \cite{elbayad2018token,chousa2018training}. These approaches use token-level and sentence-level reward functions that push the model to distribute the output probability mass over words other than the ground-truth reference. Similarly, Ma et al. \cite{ma2018bag} have added a bag-of-words term to the training objective, assuming that the set of correct translations share similar bag-of-word vectors.

There has also been extensive work on addressing the exposure bias problem. An approach that has proved effective is the incorporation of predictions in the training, via either imitation learning \cite{daume2009search,ross2011reduction,leblond2017searnn} or reinforcement learning \cite{ranzato2015sequence,bahdanau2016actor}. Another approach, that is computationally more efficient, leverages scheduled sampling to obtain a stochastic mixture of words from the reference and the predictions \cite{bengio2015scheduled}. In turn, Wu et al. \cite{xu2019differentiable} have proposed a soft alignment algorithm to alleviate the missmatches between the reference translations and the predictions obtained with scheduled sampling; and Zhang et al.\cite{zhang2018regularizing} have introduced two regularization terms based on the Kullback-Leibler (KL) divergence to improve the agreement of sentences predicted from left-to-right and right-to-left. 


\subsection{Word and Sentence Embeddings}
\label{word_and_sen_embs}

\begin{figure*}[t!]
	\centering
	\includegraphics[width=0.6\linewidth]{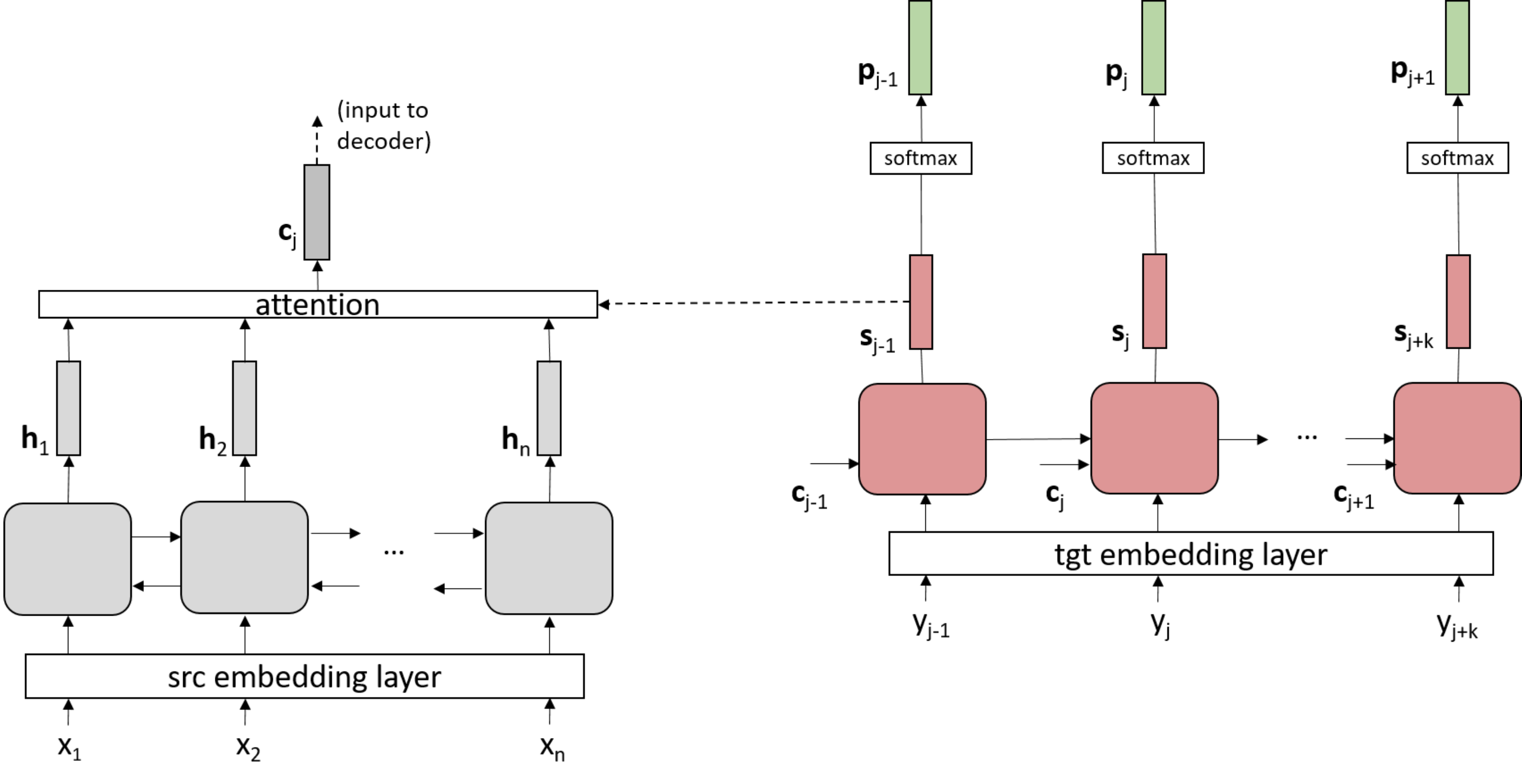}
	\caption{Baseline NMT model. (Left) The encoder receives the input sentence and generates a context vector $\textbf{c}_j$ for each decoding step using an attention mechanism. (Right) The decoder generates one-by-one the output vectors $\textbf{p}_j$, which represent the probability distribution over the target vocabulary. During training $\textbf{y}_j$ is a token from the ground truth sentence, but during inference the model uses its own predictions.}
	\label{fig:fig1}
\end{figure*}

Word vectors or word embeddings \cite{mikolov2013distributed,pennington2014glove,bojanowski2017enriching} are ubiquitous in NLP since they provide effective input features for deep learning models. Recently, contextual word vectors such as ELMo \cite{peters2018deep}, BERT \cite{devlin2018bert} and the OpenAI transformer \cite{radford2018improving} have led to remarkable performance improvements in several language understanding tasks. Additionally, researchers have focused on developing embeddings for entire sentences and documents as they may facilitate several textual classification tasks \cite{kiros2015skip,conneau2017supervised,cer2018universal,artetxe2018massively}. 

In NMT models, word embeddings play an important role as input of both the encoder and the decoder. A recent paper has shown that contextual word embeddings provide effective input features for both stages \cite{edunov2019pre}. However, very little research has been devoted to using word embeddings as targets. Kumar and Tsvetkov \cite{kumar2018mises} have removed the typical output softmax layer, forcing the decoder to generate continuous outputs. At inference time, they use a nearest-neighbour search in the word embedding space to select the word to predict. Their model allows for significantly faster training while performing on par with state-of-the-art models. Our approach differs from \cite{kumar2018mises} in that our decoder generates continuous outputs \textit{in parallel} with the standard softmax layer, and only during training to provide regularization. 
At inference time, the continuous output is ignored and prediction operates as in a standard NMT model. To the best of our knowledge, our model is the first to use embeddings as targets for regularization, and at both word and sentence level.

\subsection{Unsupervised NMT}
\label{subsec_unsupervisedNMT}

The amount of available parallel, human-annotated corpora for training NMT systems is at times very scarce. This is the case of many low-resource languages and specialized translation domains (e.g., health care). Consequently, there has been a growing interest in developing unsupervised NMT models \cite{lample2018phrase,artetxe2017unsupervised,yang2018unsupervised} which do not require annotated data for training. Such models learn to translate by only using monolingual corpora, and even though their accuracy is still well below that of their supervised counterparts, they have started to reach interesting levels. The architecture of unsupervised NMT systems differs from that of supervised systems in that it combines translation in both directions (source-to-target and target-to-source). Typically, a single encoder is used to encode sentences from both languages, and a separate decoder generates the translations in each language. The training of such systems follows three stages: 1) building a bilingual dictionary and word embedding space, 2) training two monolingual language models as denoising autoencoders \cite{vincent2008extracting}, and 3) converting the unsupervised problem into a weakly-supervised one by use of back-translations \cite{sennrich2015improving}. For more details on unsupervised NMT systems, we refer the reader to the original papers~ \cite{lample2018phrase,artetxe2017unsupervised,yang2018unsupervised}.

In this paper, we explore using the proposed regularization approach also for unsupervised NMT. Unsupervised NMT models still require very large amounts of monolingual data for training, and often such amounts are not available. Therefore, these models, too, are expected to benefit from improved regularization.  

\begin{figure*}[t!]
	\centering
	\includegraphics[width=0.7\linewidth]{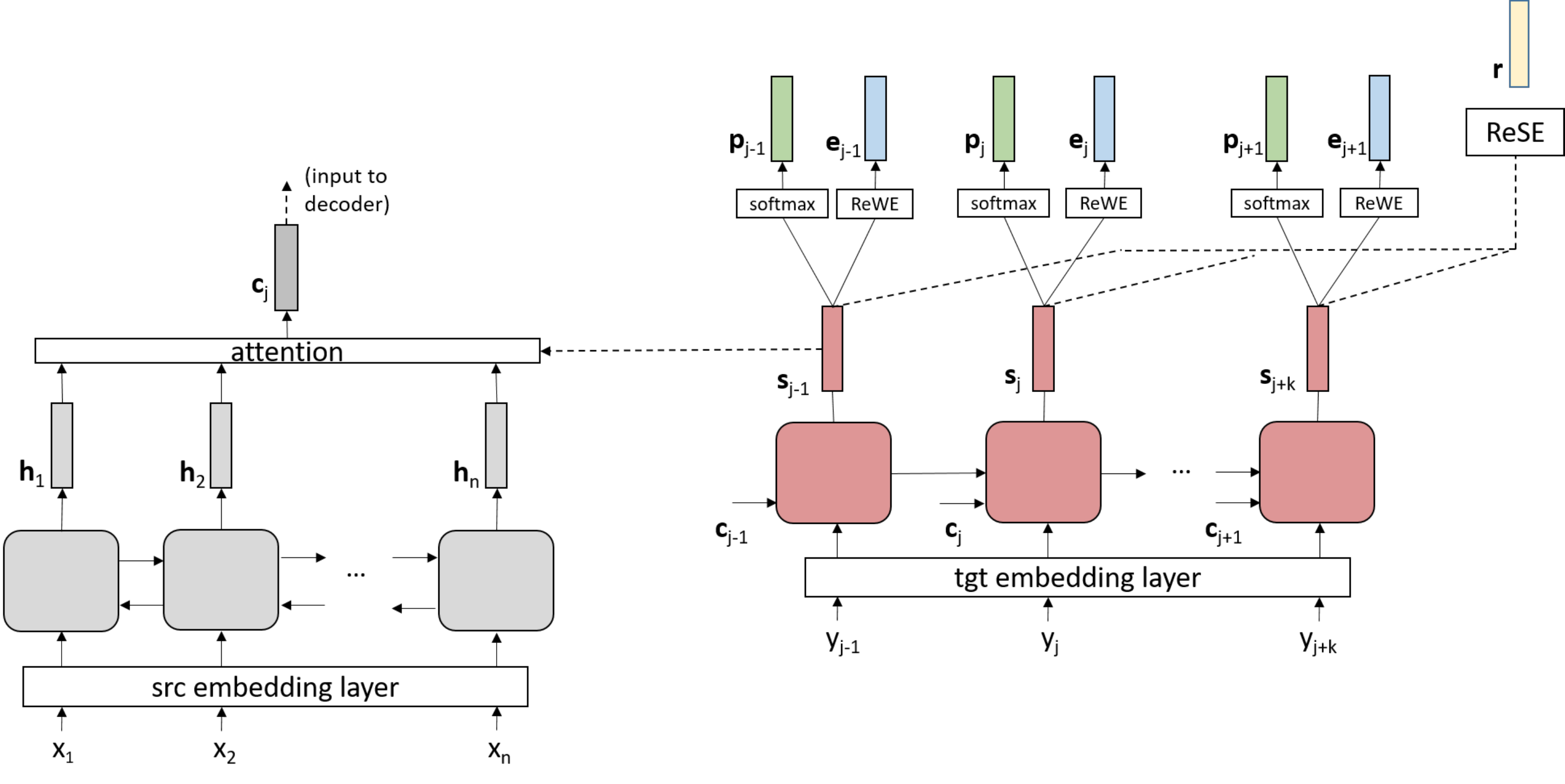}
	\caption{Full model: Baseline + ReWE + ReSE. (Left) The encoder with the attention mechanism generates vectos $\textbf{c}_j$ in the same way as the baseline system. (Right) The decoder generates one-by-one the output vectors $\textbf{p}_j$, which represent the probability distribution over the target vocabulary, and $\textbf{e}_j$, which is a continuous word vector. Additionally, the model can also generate another continuous vector, $\textbf{r}$, which represents the sentence embedding.}
	\label{fig:fig2}
\end{figure*}

\section{The Baseline NMT model}
\label{baseline_nmt_model}

In this section, we describe the NMT model that has been used as the basis for the proposed regularizer. It is a  neural encoder-decoder architecture with attention \cite{bahdanau2014} that can be regarded as a strong baseline as it incorporates both LSTMs and transformers as modules. Let us assume that $\textbf{x}:\{x_1 \dots x_n\}$ is the source sentence with $n$ tokens and $\textbf{y}:\{y_1 \dots y_m\}$ is the target translated sentence with $m$ tokens. First, the words in the source sentence are encoded into their word embeddings by an embedding layer:

\begin{equation}
\label{eq:src_emb}
\begin{split}
\textbf{x}^{e}_{i}=SrcEmbLayer(x_i) \quad i=1 \dots n 
\end{split}
\end{equation}

\noindent and then the source sentence is encoded by a sequential module into its hidden vectors, ${\textbf{h}_1 \dots \textbf{h}_n}$:

\begin{equation}
\label{eq:encoder}
\begin{split}
\textbf{h}_{i}=enc(\textbf{h}_{i-1},\textbf{x}^{e}_{i}) \quad i=1 \dots n 
\end{split}
\end{equation}

Next, for each decoding step $j=1 \ldots m$, an attention network provides a context vector $\textbf{c}_j$ as a weighted average of all the encoded vectors, $\textbf{h}_1 \dots \textbf{h}_n$, conditional on the decoder output at the previous step, $\textbf{s}_{j-1}$ (Eq. \ref{eq:attention}). For this network, we have used the attention mechanism of Badhdanau et al.\cite{bahdanau2014}.

\begin{equation}
\label{eq:attention}
\begin{split}
\textbf{c}_{j}=attn(\textbf{h}_1 \dots \textbf{h}_n,\textbf{s}_{j-1}) \quad j=1 \dots m
\end{split}
\end{equation}

Given the context vector, $\textbf{c}_j$, the decoder output at the previous step, $\textbf{s}_{j-1}$, and the word embedding of the previous word in the target sentence, $\textbf{y}^{e}_{j}$ (Eq. \ref{eq:tgt_emb}), the decoder generates vector $\textbf{s}_j$ (Eq. \ref{eq:decoder}). This vector is later transformed into a larger vector of the same size as the target vocabulary via learned parameters $\textbf{W}$, $\textbf{b}$ and a softmax layer (Eq. \ref{eq:generator}). The resulting vector, $\textbf{p}_j$, is the inferred probability distribution over the target vocabulary at decoding step $j$. Fig. \ref{fig:fig1} depicts the full architecture of the baseline model.

\begin{equation}
\label{eq:tgt_emb}
\begin{split}
\textbf{y}^{e}_{j}=TgtEmbLayer(y_j) \quad j=1 \dots m 
\end{split}
\end{equation}

\begin{equation}
\label{eq:decoder}
\begin{split}
\textbf{s}_{j}=dec(\textbf{c}_j,\textbf{s}_{j-1},\textbf{y}^{e}_{j-1}) \quad j=1 \dots m
\end{split}
\end{equation}

\begin{equation}
\label{eq:generator}
\begin{split}
\textbf{p}_{j}= softmax(\textbf{W}\textbf{s}_j+\textbf{b}) 
\end{split}
\end{equation}

The model is trained by minimizing the negative log-likelihood (NLL) which can be expressed as:

\begin{equation}
\label{eq:NLL_loss}
\begin{split}
\Lagr_{NLL} = -\sum_{j=1}^{m}\log(\textbf{p}_{j}(y_j))
\end{split}
\end{equation}

\noindent where the probability of ground-truth word ${y}_j$ has been noted as $\textbf{p}_{j}({y}_{j})$. Minimizing the NLL is equivalent to MLE and results in assigning maximum probability to the words in the reference translation, $y_j, j=1 \ldots m$. The training objective is minimized with standard backpropagation over the training data, and at inference time the model uses beam search for decoding.

\section{Regressing word and sentence embeddings}
\label{rewe_and_rese}

As mentioned in the introduction, MLE suffers from some limitations when training a neural machine translation system. To alleviate these shortcomings, in our recent paper \cite{jauregi2019rewe} we have proposed a new regularization method based on regressing word embeddings. In this paper, we extend this idea to sentence embeddings.

\subsection{ReWE}
\label{rewe}

Pre-trained word embeddings are trained on large monolingual corpora by measuring the co-occurences of words in text windows (``contexts''). Words that occur in similar contexts are assumed to have similar meaning, and hence, similar vectors in the embedding space. Our goal with ReWE is to incorporate the information embedded in the word vector in the loss function to encourage model regularization.

In order to generate continuous vector representations as outputs, we have added a ReWE block to the NMT baseline (Fig. \ref{fig:fig2}). At each decoding step, the ReWE block receives the hidden vector from the decoder, $\textbf{s}_j$, as input and outputs another vector, $\textbf{e}_j$, of the same size of the pre-trained word embeddings:

\begin{equation}
\label{eq:emb_generator}
\begin{split}
\textbf{e}_j &= ReWE(\textbf{s}_j) \\ &= \textbf{W}_2(ReLU(\textbf{W}_1\textbf{s}_j+\textbf{b}_1))+\textbf{b}_2
\end{split}
\end{equation}

\noindent where $\textbf{W}_1$, $\textbf{W}_2$, $\textbf{b}_1$ and $\textbf{b}_2$ are the learnable parameters of a two-layer feed-forward network with a Rectified Linear Unit (ReLU) as  activation function between the layers. Vector $\textbf{e}_j$ aims to reproduce the word embedding of the target word, and thus the distributional properties (or co-occurrences) of its contexts. 

During training, the model is guided to regress the predicted vector, $\textbf{e}_j$, towards the word embedding of the ground-truth word, $\textbf{y}^{e}_j$. This is achieved by using a loss function that computes the distance between $\textbf{e}_j$ and $\textbf{y}^{e}_j$ (Eq. \ref{eq:L_ReWE}). Previous work \cite{jauregi2019rewe} has showed that the cosine distance is empirically an effective distance between word embeddings and has thus been adopted as loss. This loss and the original NLL loss are combined together with a tunable hyper-parameter, $\lambda$ (Eq. \ref{eq:L_w}). Therefore, the model is trained to jointly predict both a categorical and a continuous representation of the words. Even though the system is performing a single task, this setting could also be interpreted as a form of multi-task learning with different representations of the same targets.

\begin{equation}
\label{eq:L_ReWE}
\begin{split}
\Lagr_{ReWE} = \sum_{j=1}^{m}(1 - cos(\textbf{e}_j,\textbf{y}^{e}_{j}))
\end{split}
\end{equation}

\begin{equation}
\label{eq:L_w}
\begin{split}
\Lagr_{w} = \Lagr_{NLL} + \lambda\Lagr_{ReWE}
\end{split}
\end{equation}

The word vectors of both the source ($\textbf{x}^{e}$) and target ($\textbf{y}^{e}$) vocabularies are initialized with pre-trained embeddings, but updated during training. At inference time, we ignore the outputs of the ReWE block and we perform translation using only the categorical prediction. 

\subsection{ReSE}
\label{rese}

Sentence vectors, too, have been extensively used as input representations in many NLP tasks such as text classification, paraphrase detection, natural language inference and question answering. The intuition behind them is very similar to that of word embeddings: sentences with similar meanings are expected to be close to each other in vector space. Many off-the-shelf sentence embedders are currently available and they can be easily integrated in deep learning models. Based on similar assumptions to the case of word embeddings, we have hypothesized that an NMT model could also benefit from a regularization term based on regressing sentence embeddings (the ReSE block in Fig. \ref{fig:fig2}).

The main difference of ReSE compared to ReWE is that there has to be a single regressed vector per sentence rather than one per word. Thus, ReSE first uses a self-attention mechanism to learn a weighted average of the decoder's hidden vectors, $\textbf{s}_1 \dots \textbf{s}_m$: 

\begin{equation}
\label{eq:SelfAttn}
\begin{split}
SelfAttn(\textbf{s}_1,\dots,\textbf{s}_m) = \sum_{j=0}^{m}\alpha_j\textbf{s}_j
\end{split}
\end{equation}

\begin{equation}
\label{eq:softmax}
\begin{split}
\alpha_j = \frac{e^{l_j}}{\sum_{k=0}^{m}e^{l_k}}
\end{split}
\end{equation}

\begin{equation}
\label{eq:attn_score}
\begin{split}
l_j = \textbf{U}_2\tanh(\textbf{U}_1\textbf{s}_j)
\end{split}
\end{equation}

\noindent where the $\alpha_j$ attention weights are obtained from Eqs. \ref{eq:softmax} and \ref{eq:attn_score}, and $\textbf{U}_1$ and $\textbf{U}_2$ are learnable parameters. Then, a two-layered neural network similar to ReWE's predicts the sentence vector, $\textbf{r}$ (Eq. \ref{eq:ReSE}). Parameters $\textbf{W}_3$, $\textbf{W}_4$, $\textbf{b}_3$ and $\textbf{b}_4$ are also learned during training.

\begin{equation}
\label{eq:ReSE}
\begin{split}
\textbf{r} &= ReSE([\textbf{s}_1,\dots,\textbf{s}_m]) \\ &=
\textbf{W}_3(ReLU(\textbf{W}_4 SelfAttn([\textbf{s}_1,\dots,\textbf{s}_m])+\textbf{b}_3))+\textbf{b}_4
\end{split}
\end{equation}

Similarly to ReWE, a loss function computes the cosine distance between the predicted sentence vector, $\textbf{r}$, and the sentence vector inferred with the  off-the-shelf sentence embedder, $\textbf{y}^r$ (Eq. \ref{eq:L_ReSE}). This loss is added to the previous objective as an extra term with an additional, tunable hyper-parameter, $\beta$:

\begin{equation}
\label{eq:L_ReSE}
\begin{split}
\Lagr_{ReSE} = 1 - cos(\textbf{r},\textbf{y}^{r})
\end{split}
\end{equation}

\begin{equation}
\label{eq:L_ws}
\begin{split}
\Lagr_{ws} = \Lagr_{NLL} + \lambda\Lagr_{ReWE} + \beta\Lagr_{ReSE}
\end{split}
\end{equation}

Since the number of sentences is significantly lower than that of the words, $\beta$ typically needs to be higher than $\lambda$. Nevertheless, we tune it blindly using the validation set. The reference sentence embedding, $\textbf{y}^{r}$, can be inferred with any off-the-shelf pre-trained embedder. At inference time, the model solely relies on the categorical prediction and ignores the predicted word and sentence vectors.

\section{Experiments}
\label{experiments}

We have carried out an ample range of experiments to probe the performance of the proposed regularization approaches. This section describes the datasets, the models and the hyper-parameters used, and presents and discusses all results.

\subsection{Datasets}
\label{datasets}

Four different language pairs have been selected for the experiments. The datasets' size varies from tens of thousands to millions of sentences to test the regularizers' ability to improve translation over a range of low-resource and high-resource language pairs. 

\begin{itemize}
	\item[] \textbf{De-En}: The German-English dataset (de-en) has been taken from the WMT18 news translation shared task\footnote{WMT18: http://www.statmt.org/wmt18/translation-task.html}. The training set contains over 5M sentence pairs collected from the \textit{Europarl}, \textit{CommonCrawl} and \textit{Newscommentary} parallel corpora. As validation and test sets, we have used the \textit{newstest2017} and the \textit{newstest2018} datasets, respectively. We consider this dataset as a high-resource case.
	
	\item[] \textbf{En-Fr}: The English-French dataset (en-fr) has been sourced from the IWSLT 2016 translation shared task\footnote{IWSLT16: https://workshop2016.iwslt.org/}. This corpus contains translations of TED talks of very diverse topics. The training data provided by the organizers consist of $219,777$ translations which allow us to categorize this dataset as low/medium-resource. Following Denkowski and Neubig~\cite{denkowski2017stronger}, the validation set has been formed by merging the 2013 and 2014 test sets from the same shared task, and the test set has been formed with the 2015 and 2016 test sets. 
	
	\item[] \textbf{Cs-En}: The Czech-English dataset (cs-en) is also from the IWSLT 2016 TED talks translation task. However, this dataset is approximately half the size of en-fr as its training set consists of $114,243$ sentence pairs. Again following Denkowski and Neubig \cite{denkowski2017stronger}), the validation set has been formed by merging the 2012 and 2013 test sets, and the test set by merging the 2015 and 2016 test sets. We regard this dataset as low-resource.
	
	\item[] \textbf{Eu-En}: The Basque-English dataset (eu-en) has been collected from the WMT16 IT-domain translation shared task\footnote{WMT16 IT: http://www.statmt.org/wmt16/it-translation-task.html}. This is the smallest dataset, with only $89,413$ sentence pairs in the training set. However, only $2,000$ sentences in the training set have been translated by human annotators. The remaining sentence pairs are translations of IT-domain short phrases and Wikipedia titles. Therefore, we consider this dataset as extremely low-resource. It must be said that translations in the IT domain are somehow easier than in the news domain, as this domain is very specific and the wording of the sentences are less varied. For this dataset, we have used the validation and test sets ($1,000$ sentences each) provided in the shared task.
\end{itemize}

All the datasets have been pre-processed with \textit{moses-tokenizer}\footnote{URL moses}. Additionally, words have been split into subword units using byte pair encoding (BPE) \cite{sennrich2016neura}. For the BPE merge operations parameter, we have used $32,000$ (the default value) for all the datasets, except for eu-en where we have set it to $8,000$ since this dataset is much smaller. Experiments have been performed at both word and subword level since morphologically-rich languages such as German, Czech and Basque can benefit greatly from operating the NMT model at subword level.

\subsection{Model Training and Hyper-Parameter Selection}
\label{hyper_parameter_selection}

To implement ReWE and ReSE, we have modified the popular OpenNMT open-source toolkit \cite{klein2017opennm}\footnote{Our code is publicly available on Github at: https://github.com/ijauregiCMCRC/ReWE\_and\_ReSE.git. We will also release it on CodeOcean.}. Two variants of the standard OpenNMT model have been used as baselines: the LSTM and the transformer, described hereafter.

\begin{table}[t]
	\caption{BLEU scores over the En-Fr test set. The reported results are the average of 5 independent runs.}
	\begin{center}
		\resizebox{0.4\textwidth}{!}{\begin{tabular}{|l|c|c|}
				\hline
				\textbf{Models}&Word/BPE&BLEU\\
				\hline
				LSTM &word&34.21\\
				LSTM + ReWE($\lambda=20$) &word&35.43\\
				\hline
				Transformer &word&34.56\\
				Transformer + ReWE($\lambda=20$) &word&35.3\\
				\hline
				LSTM &bpe&34.06\\
				LSTM  + ReWE($\lambda=20$) &bpe&35.09\\
				\hline
				Transformer &bpe&35.31\\
				Transformer + ReWE($\lambda=20$) &bpe&\textbf{36.3}\\
				\hline
		\end{tabular}}
		\label{tab_en_fr}
	\end{center}
\end{table}

\begin{table}[t]
	\caption{BLEU scores over the Cs-En test set. The reported results are the average of 5 independent runs.}
	\begin{center}
		\resizebox{0.4\textwidth}{!}{\begin{tabular}{|l|c|c|}
				\hline
				\textbf{Models}&Word/BPE&BLEU\\
				\hline
				LSTM &word&20.48\\
				+ ReWE($\lambda=20$) &word&21.81\\
				+ ReWE($\lambda=20$) + ReSE($\beta=100$) &word&21.98\\
				\hline
				Transformer &word&20.56\\
				+ ReWE($\lambda=20$) &word&21.16\\
				+ ReWE($\lambda=20$) + ReSE($\beta=100$) &word&20.05\\
				\hline
				LSTM &bpe&22.56\\
				+ ReWE($\lambda=20$) &bpe&\textbf{23.72}\\
				+ ReWE($\lambda=20$) + ReSE($\beta=100$) &bpe&23.56\\
				\hline
				Transformer &bpe&21.02\\
				+ ReWE($\lambda=20$) &bpe&22.19\\
				+ ReWE($\lambda=20$) + ReSE($\beta=100$) &bpe&20.53\\
				\hline
		\end{tabular}}
		\label{tab_cs_en}
	\end{center}
\end{table}

\begin{table}[t]
	\caption{BLEU scores over the Eu-En test set. The reported results are the average of 5 independent runs.}
	\begin{center}
		\resizebox{0.4\textwidth}{!}{\begin{tabular}{|l|c|c|}
				\hline
				\textbf{Models}&Word/BPE&BLEU\\
				\hline
				LSTM &word&10.87\\
				+ ReWE($\lambda=20$) &word&13.83\\
				+ ReWE($\lambda=20$) + ReSE($\beta=100$) &word&16.02\\
				\hline
				Transformer &word&12.15\\
				+ ReWE($\lambda=20$) &word&13.53\\
				+ ReWE($\lambda=20$) + ReSE($\beta=100$) &word&6.92\\
				\hline
				LSTM &bpe&17.14\\
				+ ReWE($\lambda=20$) &bpe&19.54\\
				+ ReWE($\lambda=20$) + ReSE($\beta=100$) &bpe&\textbf{20.29}\\
				\hline
				Transformer &bpe&12.70\\
				+ ReWE($\lambda=20$) &bpe&13.21\\
				+ ReWE($\lambda=20$) + ReSE($\beta=100$) &bpe&9.63\\
				\hline
		\end{tabular}}
		\label{tab_eu_en}
	\end{center}
\end{table}

\begin{table}[t]
	\caption{BLEU scores over the De-En test set. The reported results are the average of 5 independent runs.}
	\begin{center}
		\resizebox{0.4\textwidth}{!}{\begin{tabular}{|l|c|c|}
				\hline
				\textbf{Models}&Word/BPE&BLEU\\
				\hline
				LSTM &word&29.75\\
				+ ReWE($\lambda=2$) &word&30.17\\
				+ ReWE($\lambda=2$) + ReSE($\beta=2$) &word&30.23\\
				\hline
				LSTM &bpe&\textbf{34.03}\\
				+ ReWE($\lambda=2$) &bpe&33.66\\
				+ ReWE($\lambda=2$) + ReSE($\beta=2$) &bpe&33.91\\
				\hline
		\end{tabular}}
		\label{tab_de_en}
	\end{center}
\end{table}

\begin{itemize}
	\item[] \textbf{LSTM}: A strong NMT baseline was prepared by following the indications given by Denkowski and Neubig \cite{denkowski2017stronger}. The model uses a bidirectional LSTM~\cite{hochreiter1997long} for the encoder and a unidirectional LSTM for the decoder, with two layers each. The size of the word embeddings was set to 300d and that of the sentence embeddings to 512d. The sizes of the hidden vectors of both LSTMs and of the attention network were set to 1024d. In turn, the LSTM's dropout rate was set to $0.2$ and the training batch size was set to 40 sentences. As optimizer, we have used Adam \cite{kingma2014adam} with a learning rate of $0.001$. During training, the learning rate was halved with simulated annealing upon convergence of the perplexity over the validation set, which was evaluated every $25,000$ training sentences. Training was stopped after halving the learning rate $5$ times.
	
	\item[] \textbf{Transformer}: The transformer network \cite{vaswani2017attention} has somehow become the \textit{de-facto} neural network for the encoder and decoder of NMT pipelines thanks to its strong empirical accuracy and highly-parallelizable training. For this reason, we have used it as another baseline for our model. For its hyper-parameters, we have used the default values set by the developers of OpenNMT\footnote{Transformer: http://opennmt.net/OpenNMT-py/FAQ.html}. Both the encoder and the decoder are formed by a 6-layer network. The sizes of the word embeddings, the hidden vectors and the attention network have all been set to either 300d or 512d, depending on the best results over the validation set. The head count has been set correspondingly to either 6 or 8, and the dropout rate to $0.2$ as for the LSTM. The model was also optimized using Adam, but with a much higher learning rate of $1$ (OpenAI default). For this model, we have not used simulated annealing since some preliminary experiments showed that it did penalize performance. The batch size used was $4,096$ and $1,024$ words, again selected based on the accuracy over the validation set. Training was stopped upon convergence in perplexity over the validation set, which was evaluated at every epoch.
\end{itemize}

In addition, the word embeddings for both models were initialized with pre-trained fastText embeddings~\cite{bojanowski2017enriching}. For the 300d word embeddings, we have used the word embeddings available on the official fastText website\footnote{Fasttext: https://fasttext.cc/docs/en/crawl-vectors.html}. For the 512d embeddings and the subword units, we have trained our own pre-trained vectors using the fastText embedder with a large monolingual corpora from Wikipedia\footnote{Wikipedia: https://linguatools.org/tools/corpora/} and the training data. Both models have used the same sentence embeddings which have been computed with the Universal Sentence Encoder (USE)\footnote{USE: https://tfhub.dev/google/universal-sentence-encoder/2}. However, the USE is only available for English, so we have only been able to use ReSE with the datasets where English is the target language (i.e., de-en, cs-en and eu-en). When using BPE, the subwords of every sentence have been merged back into words before passing them to the USE. The BLEU score for the BPE models has also been computed after post-processing the subwords back into words. Finally, hyper-parameters $\lambda$ and $\beta$ have been tuned only once for all datasets by using the en-fr validation set. This was done in order to save the significant computational time that would have been required by further hyper-parameter exploration. However, in the de-en case the initial results were far from the state of the art and we therefore repeated the selection with its own validation set. For all experiments, we have used an Intel Xeon E5-2680 v4 with an NVidia GPU card Quadro P5000. On this machine, the training time of the transformer has been approximately an order of magnitude larger than that of the LSTM.

\begin{table}[t!]
	\caption{Translation examples. Example 1: Eu-En and Example 2: Cs-En.}\label{example_1}
	\centering
	\resizebox{0.45\textwidth}{!}{\begin{tabularx}{0.9\columnwidth}{|l X|}
			\hline
			\textbf{Example 1}: &\\
			\hline
			\textbf{Src}: &Sakatu Fitxategia fitxa Oihal atzeko ikuspegia atzitzeko ; sakatu Berria . Hautatu txantiloia eta sakatu Sortu hautatutako txantiloia erabiltzeko .\\
			\hline
			\textbf{Ref}: & Click the File tab to access Backstage view , select New . Select a template and click Create to use the selected template .\\
			\hline
			\textbf{Baseline}: &Click the default tab of the tab that you want to open the tab tab . Select the template and select the selected template .\\
			\hline
			\textbf{Baseline + ReWE}: &Press the File tab to access the view view ; click New . Select the template and click Add to create the selected template .\\
			\hline
			\textbf{Baseline + ReWE + ReSE}: &Press the File tab to access the chart view ; press New . Select the template and click Create to use the selected template .\\
			\hline
	\end{tabularx}}
\end{table}

\begin{table}[t!]
	\centering
	\resizebox{0.45\textwidth}{!}{\begin{tabularx}{0.9\columnwidth}{|l X|}
			\hline
			\textbf{Example 2}: &\\
			\hline
			\textbf{Src}: &Na tomto projektu bylo skvělé , že žáci viděli lokální problém a bum – okamžitě se s ním snaží vyrovnat .\\
			\hline
			\textbf{Ref}: &What was really cool about this project was that the students saw a local problem , and boom – they are trying to immediately address it .\\
			\hline
			\textbf{Baseline}: &In this project ,  it was great that the kids had seen local problems and boom – immediately he’s trying to deal with him .\\
			\hline
			\textbf{Baseline + ReWE}: &In this project , it was great that the kids saw a local issue ,  and boom – they immediately try to deal with it .\\
			\hline
			\textbf{Baseline + ReWE + ReSE}: &What was great about this project was that the students saw a local problem, and boom , they’re trying to deal with him .\\
			\hline
	\end{tabularx}}
	\label{example_2}
\end{table}

\subsection{Results}
\label{results_with_supervised_training}

We have carried out a number of experiments with both baselines. The scores reported are an average of the BLEU scores (in percentage points, or pp)~\cite{papineni2002bleus} over the test sets of $5$ independently trained models. Table \ref{tab_en_fr} shows the results over the en-fr dataset. In this case, the models with ReWE have outperformed the LSTM and transformer baselines consistently. The LSTM did not benefit from using BPE, but the transformer+ReWE with BPE reached $36.30$ BLEU pp (a $+0.99$ pp improvement over the best model without ReWE). For this dataset we did not use ReSE because French was the target language. 

Table \ref{tab_cs_en} reports the results over the cs-en dataset. Also in this case, all the models with ReWE have improved over the corresponding baselines. The LSTM+ReWE has achieved the best results ($23.72$ BLEU pp; an improvement of $+1.16$ pp over the best model without ReWE). This language pair has also benefited more from the BPE pre-processing, likely because Czech is a morphologically-rich language. For this dataset, it was possible to use ReSE in combination with ReWE, with an improvement for the LSTM at word level ($+0.14$ BLEU pp), but not for the remaining cases. 
We had also initially tried to use ReSE without ReWE (i.e., $\lambda=0$), but the results were not encouraging and we did not continue with this line of experiments.

\begin{figure}[t]
	\centering
	\includegraphics[width=\linewidth]{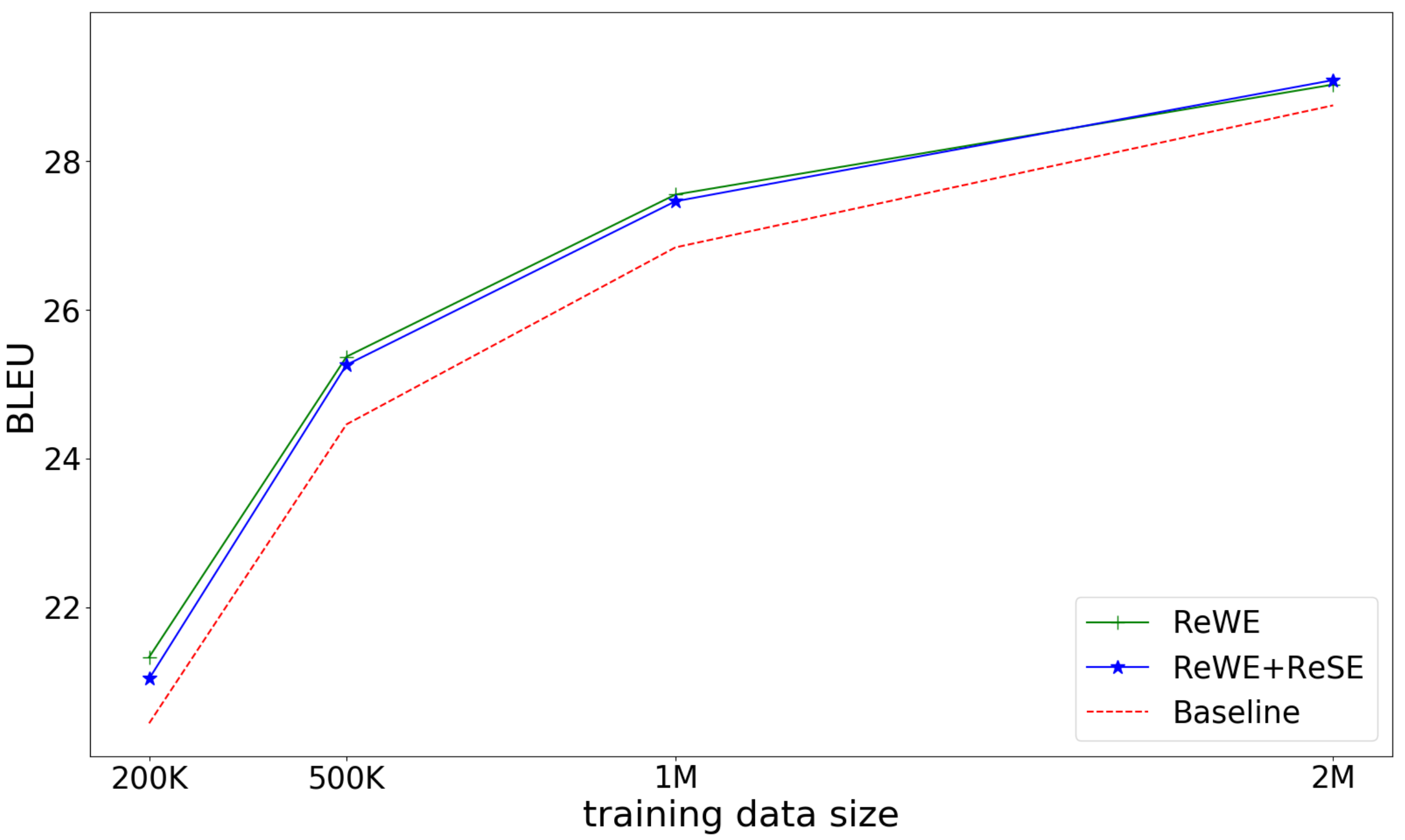}
	\caption{BLEU scores over the De-En test set for models trained with training sets of different size.}
	\label{fig:fig3}
\end{figure} 

\begin{figure*}
	\centering
	\begin{subfigure}{.8\textwidth}
		\centering
		\includegraphics[width=\linewidth]{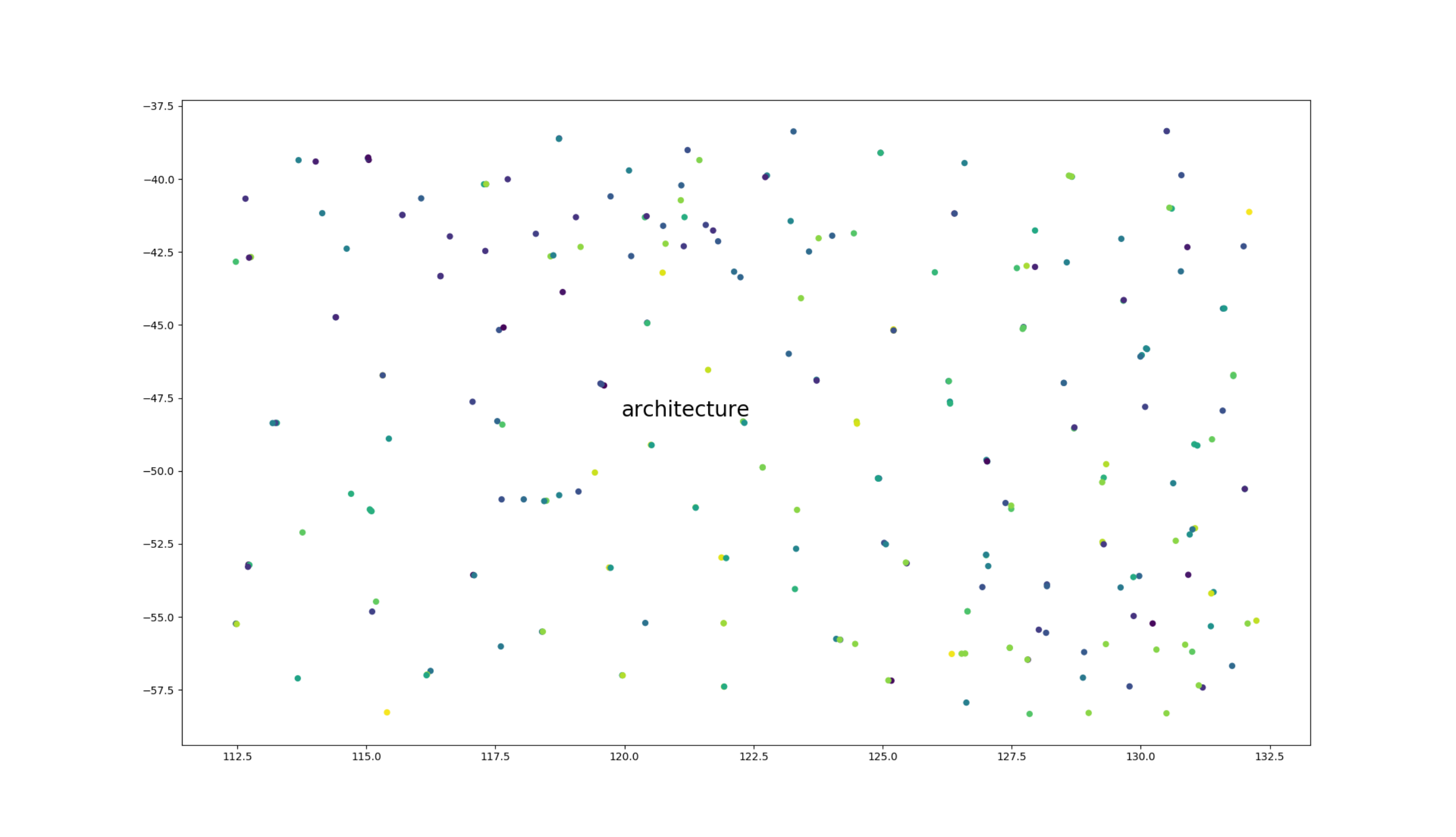}
		\caption{Baseline}
		\label{fig:sub1}
	\end{subfigure}%
	\quad
	\begin{subfigure}{.8\textwidth}
		\centering
		\includegraphics[width=\linewidth]{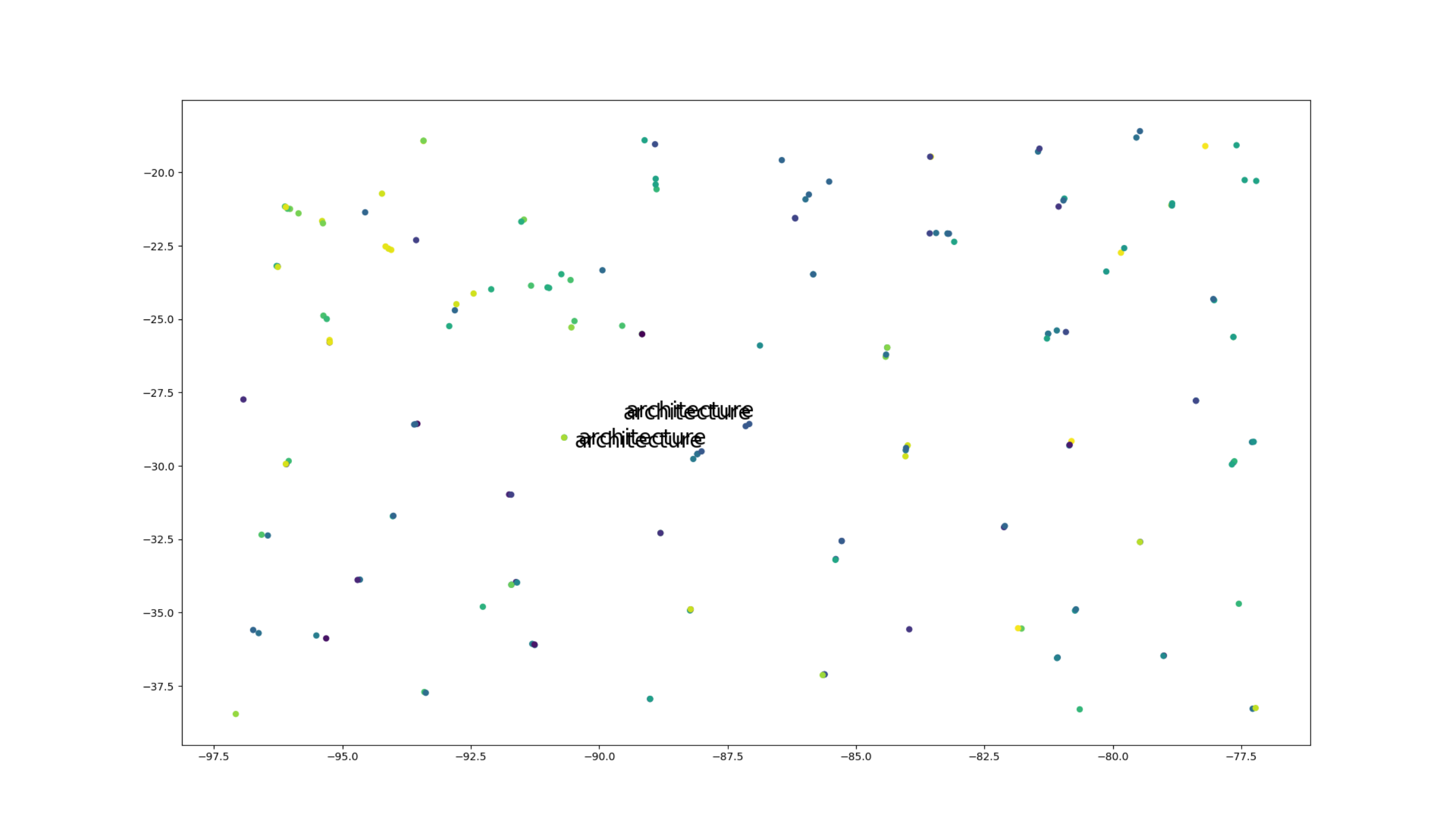}
		\caption{Baseline + ReWE}
		\label{fig:sub2}
	\end{subfigure}
	\caption{Visualization of the $\textbf{s}_j$ vectors from the decoder for a subset of the cs-en test set. Please refer to Section \ref{understanding_rewe_and_rese} for explanations. This figure should be viewed in color.}
	\label{fig:vis_1}
\end{figure*}

For the eu-en dataset (Table \ref{tab_eu_en}), the results show that, again, ReWE outperforms the baselines by a large margin. Moreover, ReWE+ReSE has been able to improve the results even further ($+3.15$ BLEU pp when using BPE and $+5.15$ BLEU pp at word level over the corresponding baselines). Basque is, too, a morphologically-rich language and using BPE has proved very beneficial ($+4.27$ BLEU pp over the best word-level model). As noted before, the eu-en dataset is very low-resource (less than $100,000$ sentence pairs) and it is more likely that  the baseline models generalize poorly. Consequently, regularizers such as ReWE and ReSE are more helpful, with larger margins of improvement with respect to the baselines. On a separate note, the transformer has unexpectedly performed well below the LSTM on this dataset, and especially so with BPE. We speculate that it may be more sensitive than the LSTM to the dataset's much smaller size, or in need of more refined hyper-parameter tuning.

Finally, Table \ref{tab_de_en} shows the results over the de-en dataset that we categorize as high-resource (5M+ sentence pairs). For this dataset, we have only been able to perform experiments with the LSTM due to the exceedingly long training times of the transformer. At word level, both ReWE and ReWE+ReSE have been able to outperform the baseline, although the margins of improvement have been smaller than for the other language pairs ($+0.42$ and $+0.48$ BLEU pp, respectively). However, when using BPE both ReWE and ReWE+ReSE have performed slightly below the baseline ($-0.37$ and $-0.12$ points BLEU pp, respectively). This shows that when the training data are abundant, ReWE or ReSE may not be beneficial. To probe this further, we have repeated these experiments by training the models over subsets of the training set of increasing size (200K, 500K, 1M, and 2M sentence pairs). Fig. \ref{fig:fig3} shows the BLEU scores achieved by the baseline and the regularized models for the different training data sizes. The plot clearly shows that the performance margin increases as the training data size decreases, as expected from a regularized model.

Table \ref{example_1} shows two examples of the translations made by the different LSTM models for eu-en and cs-en. A qualitative analysis of these examples shows that both ReWE and ReWE+ReSE have improved the quality of these translations. In the eu-en example, ReWE has correctly translated ``File tab''; and ReSE has correctly added ``click Create''. In the cs-en example, the model with ReWE has picked the correct subject ``they'', and only the model with ReWE and ReSE has correctly translated ``students'' and captured the opening phrase ``What was\dots about this\dots''.

\subsection{Understanding ReWE and ReSE}
\label{understanding_rewe_and_rese}

The quantitative experiments have proven that ReWE and ReSE can act as effective regularizers for low- and medium-resource NMT. Yet, it would be very interesting to understand how do they influence the training to achieve improved models. For that purpose, we have conducted an exploration of the values of the hidden vectors on the decoder end ($\textbf{s}_j$, Eq. \ref{eq:decoder}). These values are the ``feature space'' used by the final classification block (a linear transformation and a softmax) to generate the class probabilities and can provide insights on the model. For this reason, we have considered the cs-en test set and stored all the $\textbf{s}_j$ vectors with their respective word predictions. Then, we have used t-SNE \cite{maaten2008tsne} to reduce the dimensionality of the $\textbf{s}_j$ vectors to a visualizable 2d. Finally, we have chosen a particular word (\textit{architecture}) as the center of the visualization, and plotted all the vectors within a chosen neighborhood of this center word (Fig. \ref{fig:vis_1}). To avoid cluttering the figure, we have not superimposed the predicted words to the vectors, but only used a different color for each distinct word. The center word in the two subfigures (a: baseline; b: baseline+ReWE) is the same (\textit{architecture}) and from the same source sentence, so the visualized regions are comparable. The visualizations also display all other predicted instances of word \textit{architecture} in the neighborhood.

These visualizations show two interesting behaviors: 1) from eye judgment, the points predicted by the ReWE model seem more uniformly spread out; 2) instances of the same words have $\textbf{s}_j$ vectors that are close to each other. For instance, several instances of word \textit{architecture} are close to each other in Fig. \ref{fig:sub2} while a single instance appears in Fig. \ref{fig:sub2}. The overall observation is that the ReWE regularizer leads to a vector space that is easier to discriminate, i.e. find class boundaries for, facilitating the final word prediction. In order to confirm this observation, we have computed various clustering indexes over the clusters formed by the vectors with identical predicted word. As indexes, we have used the silhouette and the Davies-Bouldin indexes that are two well-known unsupervised metrics for clustering. The silhouette index ranges from -1 to +1, where values closer to 1 mean that the clusters are compact and well separated. The Davies-Bouldin index is an unbounded nonnegative value, with values closer to 0 meaning better clustering. 
Table \ref{clus_index} shows the values of these clustering indexes over the entire cs-en test set for the LSTM models. As the table shows, the models with ReWE and ReWE+ReSE have reported the best values. This confirms that applying ReWE and ReSE has a positive impact on the decoder's hidden space, ultimately justifying the increase in word classification accuracy.

\begin{table}[t]
	\caption{Clustering indexes of the LSTM models over the cs-en test set. The reported results are the average of 5 independent runs.}
	\begin{center}
		\resizebox{0.4\textwidth}{!}{\begin{tabular}{|l|c|c|}
				\hline
				\textbf{Model}&Sillhouette&Davies-Bouldin\\
				\hline
				LSTM &-0.19&1.87\\
				+ ReWE($\lambda=2$) &-0.17&\textbf{1.80}\\
				+ ReWE($\lambda=2$) + ReSE($\beta=2$) &\textbf{-0.16}&\textbf{1.80}\\
				\hline
		\end{tabular}}
		\label{clus_index}
	\end{center}
\end{table}

\begin{figure}
	\centering
	\begin{subfigure}{.5\textwidth}
		\centering
		\includegraphics[width=\linewidth]{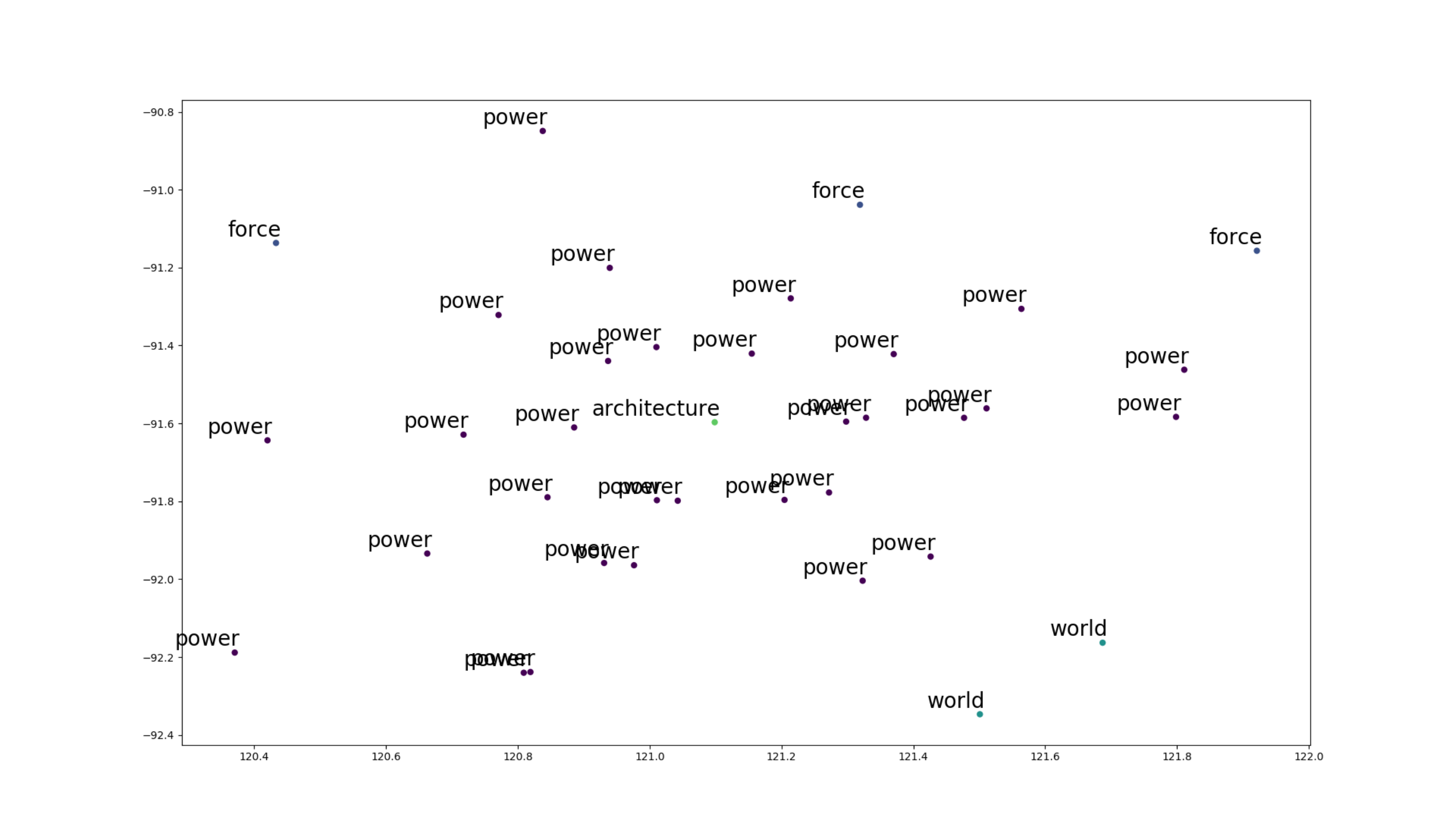}
		\caption{Baseline}
		\label{fig:sub3}
	\end{subfigure}%
	\quad
	\begin{subfigure}{.5\textwidth}
		\centering
		\includegraphics[width=\linewidth]{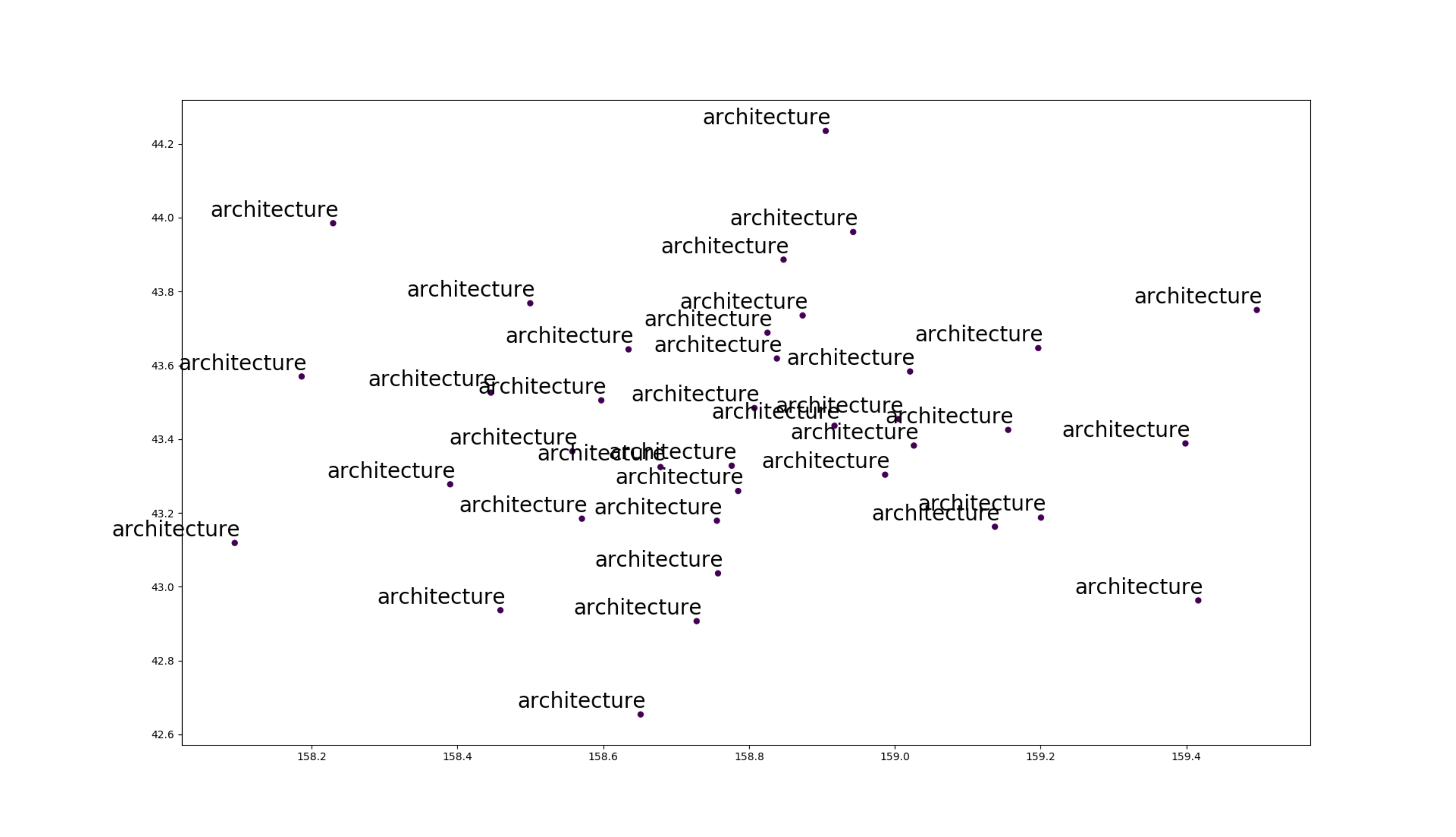}
		\caption{Baseline + ReWE}
		\label{fig:sub4}
	\end{subfigure}
	\caption{Visualization of the $\textbf{s}_j$ vectors in a smaller neighborhood of the center word.}
	\label{fig:vis_2}
\end{figure}

For further exploration, we have created another visualization of the $\textbf{s}$ vectors and their predictions over a smaller neighborhood (Fig. \ref{fig:vis_2}). The same word (\textit{architecture}) has been used as the center word of the plot. Then, we have ``vibrated'' each of  the $\textbf{s}_j$ vector by small increments (between 0.05 and 8 units) in each of their dimensions, creating several new synthetic instances of $\textbf{s}$ vectors which are very close to the original ones. These synthetic vectors have then been decoded with the trained NMT model to obtain their predicted words. Finally, we have used t-SNE to reduce the dimensionality to 2d, and visualized all the vectors and their predictions in a small neighborhood ($\pm10$ units) around the center word. Fig. \ref{fig:vis_2} shows that, with the ReWE model, all the $\textbf{s}$ vectors surrounding the center word predict the same word (\textit{architecture}). Conversely, with the baseline, the surrounding points predict different words (\textit{power}, \textit{force}, \textit{world}). This is additional evidence that the $\textbf{s}$ space is evened out by the use of the proposed regularizer.



\subsection{Unsupervised NMT}
\label{results_with_unsupervised_training}

\begin{figure*}[t]
	\centering
	\begin{subfigure}{.5\textwidth}
		\centering
		\includegraphics[width=\linewidth]{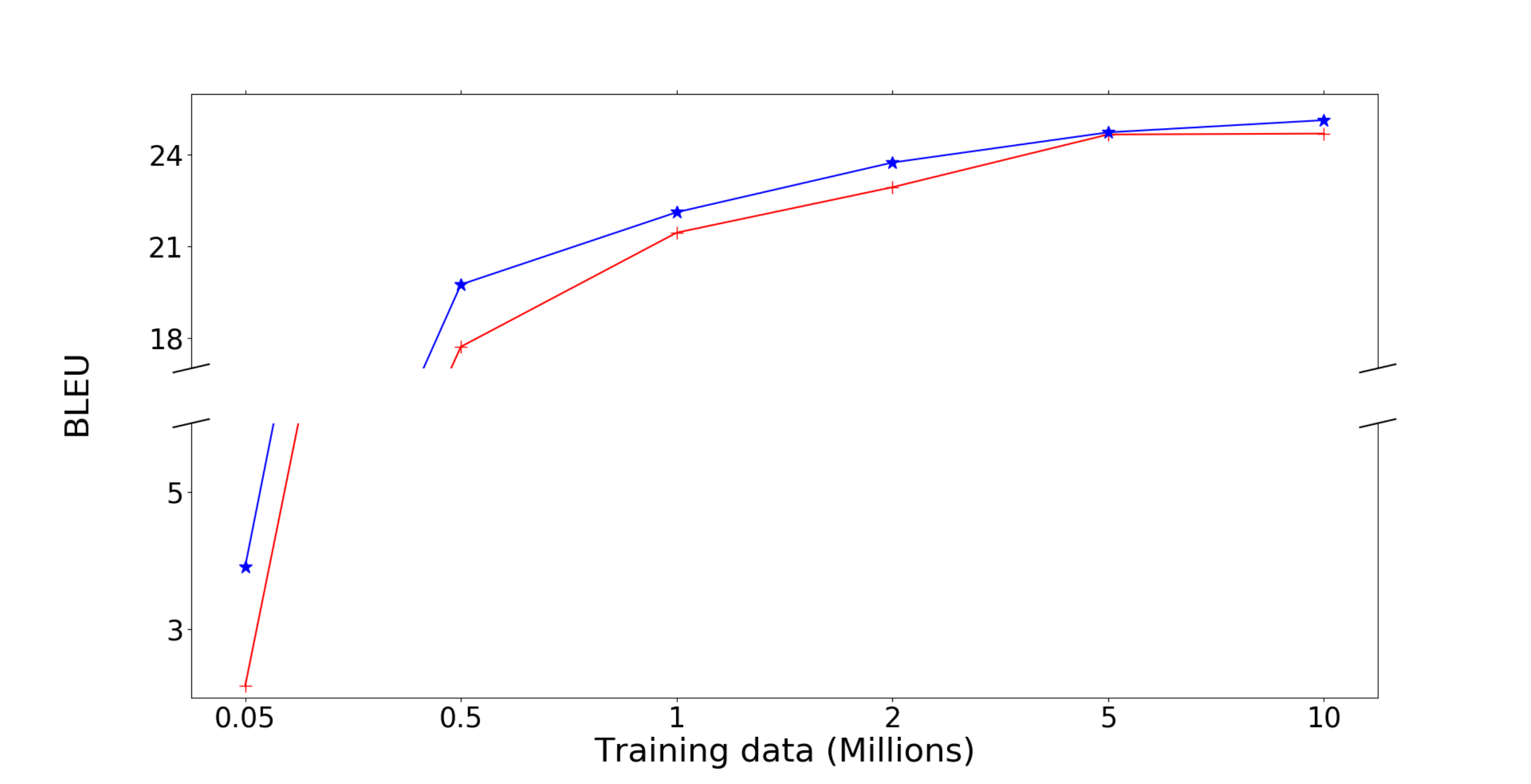}
		\caption{en-fr}
		\label{fig:sub5}
	\end{subfigure}%
	\begin{subfigure}{.5\textwidth}
		\centering
		\includegraphics[width=\linewidth]{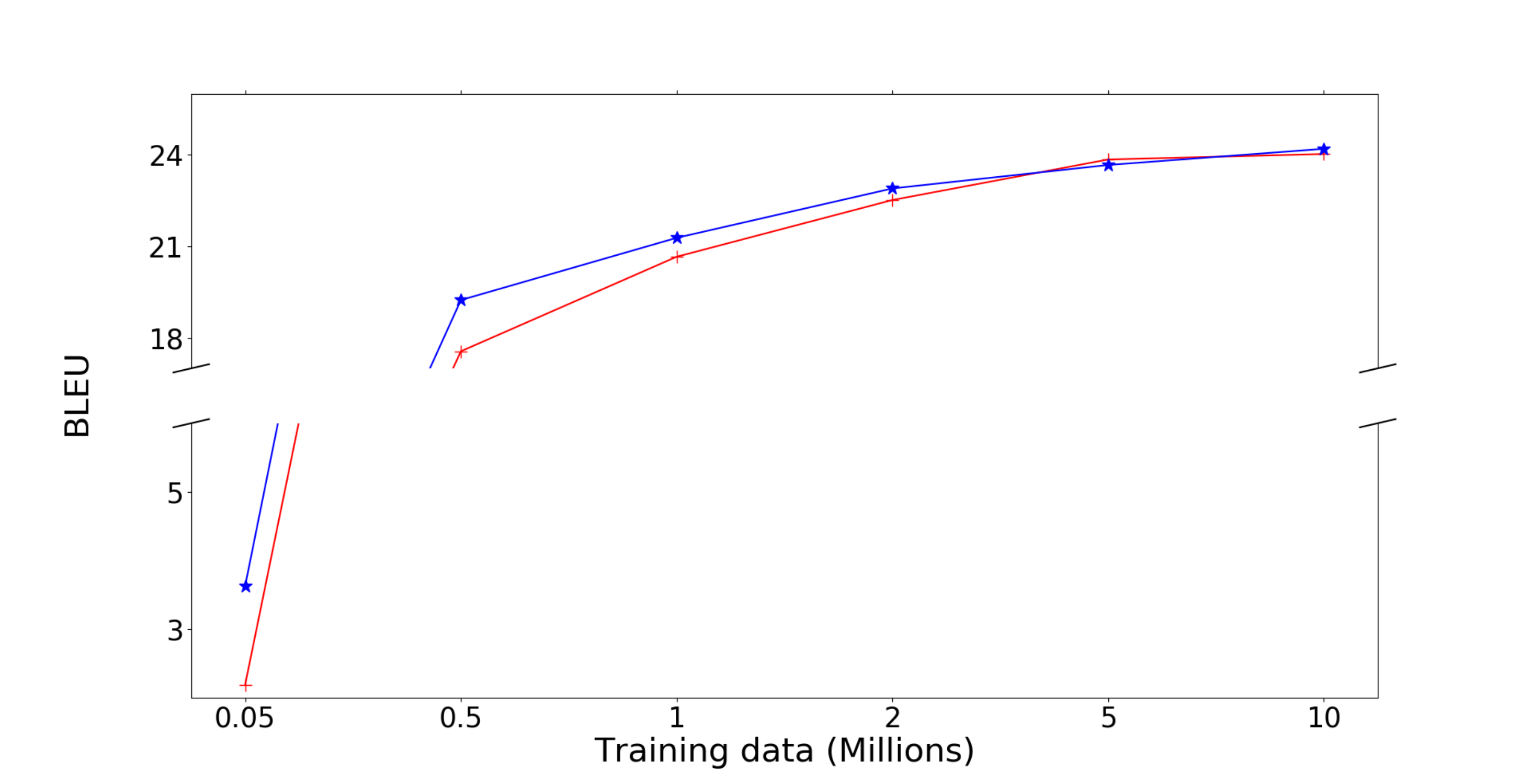}
		\caption{fr-en}
		\label{fig:sub6}
	\end{subfigure}
	\caption{BLEU scores over the test set. The reported results are the average of 5 independent runs.. The red line represents the baseline model and the blue line is the baseline + ReWE.}
	\label{fig:unsuper}
\end{figure*}

Finally, we have also experimented with the use of ReWE and ReWE+ReSE for an unsupervised NMT task. For this experiment, we have used the open-source model provided by Lample et al. \cite{lample2018phrase}\footnote{UnsupervisedMT: https://github.com/facebookresearch/UnsupervisedMT} which is currently the state of the art for unsupervised NMT, and also adopted its default hyper-parameters and pre-processing steps which include 4-layer transformers for the encoder and both decoders, and BPE subword learning. The experiments have been performed using the WMT14 English-French test set for testing in both language directions (en-fr and fr-en), and the monolingual data from that year's shared task for training. 

As described in Section \ref{subsec_unsupervisedNMT}, an unsupervised NMT model contains two decoders to be able to translate into both languages. The model is trained by iterating over two alternate steps: 1) training using the decoders as monolingual, de-noising language models (e.g., en-en, fr-fr), and 2) training using back-translations (e.g., en-fr-en, fr-en-fr). Each step requires an objective function, which is usually an NLL loss. Moreover, each step is performed in both directions (en$\rightarrow$fr and fr$\rightarrow$en), which means that an unsupervised NMT model uses a total of four different objective functions. Potentially, the regularizers could be applied to each of them. However, the pre-trained USE sentence embeddings are only available in English, not in French, and for this reason we have limited our experiments to ReWE alone. In addition, the initial results have showed that ReWE is actually detrimental in the de-noising language model step, so we have limited its use to both language directions in the back-translation step, with the hyper-parameter, $\lambda$, tuned over the validation set ($\lambda=0.2$).

To probe the effectiveness of the regularized model, Fig. \ref{fig:unsuper} shows the results over the test set from the different models trained with increasing amounts of monolingual data (50K, 500K, 1M, 2M, 5M and 10M sentences in each language). The model trained using ReWE has been able to consistently outperform the baseline in both language directions. The trend we had observed in the supervised case has applied to these experiments, too: the performance margin has been larger for smaller training data sizes. For example, in the en-fr direction the margin has been $+1.74$ BLEU points with 50K training sentences, but it has reduced to $+0.44$ BLEU points when training with 10M sentences. Again, this behavior is in line with the regularizing nature of the proposed regressive objectives.

\section{Conclusion}
\label{conclusion}

In this paper, we have proposed regressing continuous representations of words and sentences (ReWE and ReSE, respectively) as novel regularization techniques for improving the generalization of NMT models. Extensive experiments over four different language pairs of different training data size (from 89K to 5M sentence pairs) have shown that both ReWE and ReWE+ReSE have improved the performance of NMT models, particularly in low- and medium-resource cases, for increases in BLEU score up to $5.15$ percentage points. In addition, we have presented a detailed analysis showing how the proposed regularization modifies the decoder's output space, enhancing the clustering of the vectors associated with unique words. Finally, we have showed that the regularized models have also outperformed the baselines in experiments on unsupervised NMT. As future work, we plan to explore how the categorical and continuous predictions from our model could be jointly utilized to further improve the quality of the translations.

\appendices
\section*{Acknowledgment}

The authors would like to thank the RoZetta Institute (formerly CMCRC) for providing financial support to this research.

\ifCLASSOPTIONcaptionsoff
  \newpage
\fi



%
\bibliographystyle{ieeetran}
\bibliography{bibliography}

%

\begin{IEEEbiography}[{\includegraphics[width=1in,height=1.25in,clip,keepaspectratio]{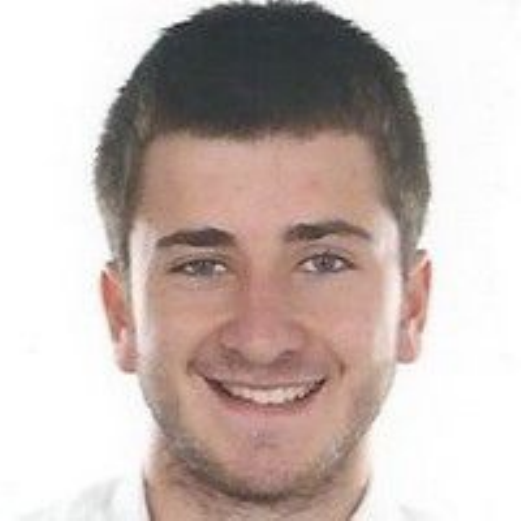}}]{Inigo Jauregi Unanue}
received the BEng degree in telecommunication systems from University of Navarra, Donostia-San Sebastian, Spain, in 2016. From 2014 to 2016, he was a research assistant at Centro de Estudio e Investigaciones Tecnicas (CEIT). Since 2016, he is a natural language processing and machine learning researcher at the RoZetta Institute (former CMCRC) in Sydney, Australia. Additionally, he is in the last year of his PhD at University of Technology Sydney, Australia. His research interests are machine learning, natural language processing and information theory.
\end{IEEEbiography}

\vspace{-10cm}

\begin{IEEEbiography}[{\includegraphics[width=1in,height=1.25in,clip,keepaspectratio]{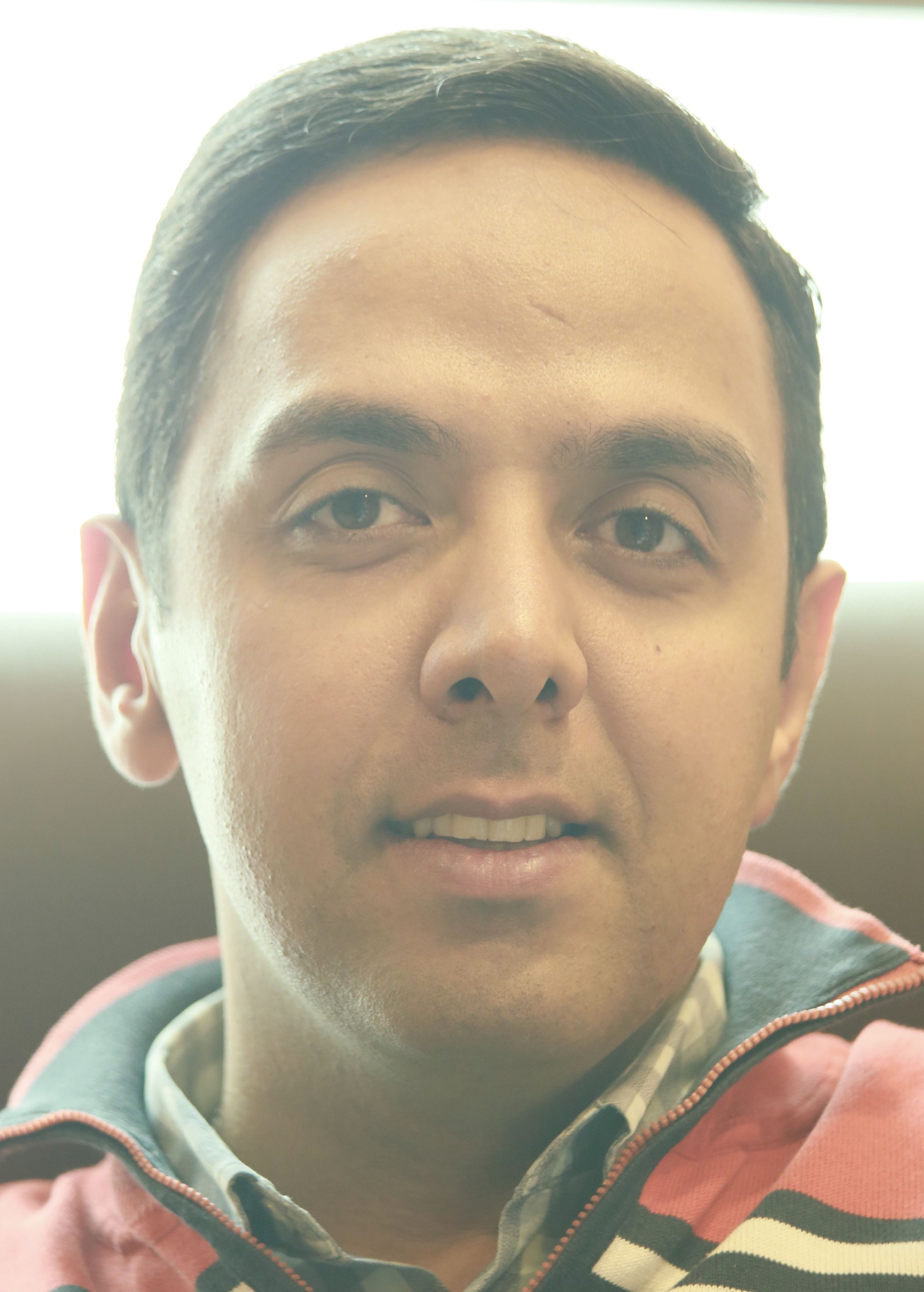}}]{Ehsan Zare Borzeshi}
received the PhD degree from University of Technology Sydney, Australia, in 2013.
He is currently a Senior Data \& Applied Scientist with Microsoft CSE (Commercial Software Engineering). He has previously held appointments as a senior researcher at the University of Newcastle,  University of Technology Sydney, and the RoZetta Institute (formerly CMCRC) in Sydney. He has also been a Visiting Scholar with the University of Central Florida, Orlando, FL, USA. His current research interests include big data, deep learning and natural language processing where he has many publications.
\end{IEEEbiography}


\vspace{-10cm}

\begin{IEEEbiography}[{\includegraphics[width=1in,height=1.25in,clip,keepaspectratio]{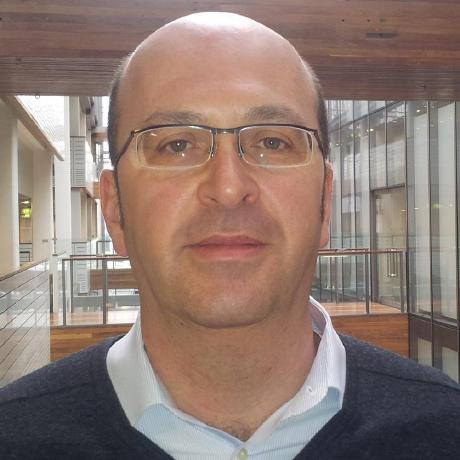}}]{Massimo Piccardi} (SM'05) received the MEng and PhD degrees from the University of Bologna, Bologna, Italy, in 1991 and 1995, respectively. He is currently a Full Professor of computer systems with University of Technology Sydney, Australia. His research interests include natural language processing, computer vision and pattern recognition and he has co-authored over 150 papers in these areas. Prof. Piccardi is a Senior Member of the IEEE, a member of its Computer and Systems, Man, and Cybernetics Societies, and a member of the International Association for Pattern Recognition. He presently serves as an Associate Editor for the IEEE Transactions on Big Data.
\end{IEEEbiography}




\end{document}